\documentclass[runningheads]{llncs}
\usepackage{graphics}
\usepackage{graphicx}
\usepackage{hyperref}
\usepackage{makecell}
\usepackage{url}
\usepackage{cite}
\usepackage{tikz}
\usetikzlibrary{shapes,snakes}
\usepackage{dirtytalk}
\usepackage{listings}
\usepackage{amsmath,amsfonts,amssymb,amscd}
\usepackage{color}
\usepackage{pgfplots}
\usepgfplotslibrary{groupplots}
\pgfplotsset{compat=newest}
\usepackage{csquotes}

\usepackage{mathptmx}
\usepackage{fancyhdr}

\usetikzlibrary{matrix,fit,positioning,arrows.meta}

\newcommand{{\modelname}}{IDNE}

\begin{document}
\title{Inductive Document Network Embedding with Topic-Word Attention}

%
%
\author{Robin Brochier\inst{1,2}\orcidID{0000-0002-6188-6509} \and Adrien Guille\inst{1}\orcidID{0000-0002-1274-6040} \and Julien Velcin\inst{1}\orcidID{0000-0002-2262-045X}}
\authorrunning{R. Brochier et al.}
%
\institute{Universit\'e de Lyon, Lyon 2, ERIC EA3083 \and Digital Scientific Research Technology \email{\{robin.brochier,adrien.guille,julien.velcin\}@univ-lyon2.fr}}
%

\cfoot{42nd European Conference on IR Research, ECIR 2020}

\maketitle              

\begin{abstract}
Document network embedding aims at learning representations for a structured text corpus \textit{i.e.} when documents are linked to each other. Recent algorithms extend network embedding approaches by incorporating the text content associated with the nodes in their formulations. In most cases, it is hard to interpret the learned representations. Moreover, little importance is given to the generalization to new documents that are not observed within the network. In this paper, we propose an interpretable and inductive document network embedding method. 
We introduce a novel mechanism, the Topic-Word Attention (TWA), that generates document representations based on the interplay between word and topic representations. We train these word and topic vectors through our general model, Inductive Document Network Embedding (IDNE), by leveraging the connections in the document network. Quantitative evaluations show that our approach achieves state-of-the-art performance on various networks and we qualitatively show that our model produces meaningful and interpretable representations of the words, topics and documents.

\end{abstract}
\section{Introduction}

Document networks, \textit{e.g.} social media, question-and-answer websites, the scientific literature, are ubiquitous. Because these networks keep growing larger and larger, navigating efficiently through them becomes increasingly difficult. 
Modern information retrieval systems rely on machine learning algorithms to support users. The performance of these systems heavily depends on the quality of the document representations. Learning good features for documents is still challenging, in particular when they are structured in a network.

Recent methods learn the representations in an unsupervised manner by combining structural and textual information. 
Text-Associated DeepWalk (TADW) \cite{yang2015network} incorporates text features into the low-rank factorization of a matrix describing the network. Graph2Gauss \cite{bojchevski2018deep} learns a deep encoder, guided by the network, that maps the nodes' attributes to embeddings. GVNR-t \cite{brochier2019global} factorizes a random walk based matrix of node co-occurrences and integrates word vectors of the documents in its formulation. CANE \cite{tu2017cane} introduces a mutual attention mechanism that builds representations of a document contextually to each of its direct neighbors in the network.           

Apart from Graph2gauss, these methods are not intended to generate representations for documents with no connection to other documents and thus cannot induce \textit{a posteriori} representations for new documents. Moreover, they provide little to no possibility to interpret the learned representations. CANE is a notable exception since its attention mechanism produces interpretable weights that highlight the words explaining the links between documents. Nevertheless, it lacks the ability to explain the representations for each document independently.     

In this paper, we describe and evaluate an inductive and interpretable method that learns word, topic and document representations in a single vector space, based on a new attention mechanism. Our contributions are the following:

\begin{itemize}
    \item we present a novel attention mechanism, Topic-Word Attention (TWA), that produces representations of a text where latent topic vectors attend to the word vectors of a document; 
    \item we explain how to train the parameters of TWA by leveraging the links of the network. Our method, Inductive Document Network Embedding (\modelname{}), is able to produce representations for previously unseen documents, without network information;
    \item we quantitatively assess the performance of \modelname{} on several networks and show that our method performs better than recent methods in various settings, including when new documents, not part of the network, are inductively represented by the algorithms. To our knowledge, we are the first to evaluate this kind of inductive setting in the context of document network embedding;
    \item we qualitatively show that our model learns meaningful word and topic vectors and produces interpretable document representations.  
\end{itemize}

The rest of the paper is organized as follows. In Section \ref{sec:related-works} we survey related works. We present in details our attention mechanism and show how to train it on networks of documents in Section \ref{sec:method}. Next, in Section \ref{sec:quanti-eval}, we present a thorough experimental study, where we assess the performance of our model following the usual evaluation protocol on node classification and further evaluating its capacity of inducting representations for text documents with no connection to the network. In Section \ref{sec:quali-eval}, we study the ability of our method to provide interpretable representations. Lastly, we conclude this paper and provide future directions in Section \ref{sec:conclusion}. The code for our model, the datasets and the evaluation procedure are made publicly available \footnote{https://github.com/brochier/idne}.

\section{Related Work}\label{sec:related-works}

Network embedding (NE) provides an efficient approach to represent nodes in a low dimensional vector space, suitable for solving various machine learning tasks. Recent techniques extend NE for document networks, showing that text and graph information can be combined to improve the resolution of classification and prediction tasks. In this section, we first cover important works in document NE and then relate recent advances in attention mechanisms.

\subsection{Document Network Embedding}

DeepWalk \cite{perozzi2014deepwalk} and node2vec \cite{grover2016node2vec} are the most well-known NE algorithms. They train dense embedding vectors by predicting nodes co-occurrences through random walks by adapting the Skip-Gram model initially designed for word embedding \cite{mikolov2013distributed}. VERSE \cite{tsitsulin2018verse} propose an efficient algorithm that can handle any type of similarity over the nodes.

Text-Associated DeepWalk (TADW) \cite{yang2015network} extends DeepWalk to deal with textual attributes. Yang \textit{et al.} prove, following the work in \cite{levy2014neural}, that Skip-Gram with hierarchical softmax can be equivalently formulated as a matrix factorization problem. TADW then consists in constraining the factorization problem with a pre-computed representation of the documents $T$ by using Latent Semantic Analysis (LSA) \cite{deerwester1990ilsa}. The task is to optimize the objective: 
\begin{align} \label{eq:tadw}
\mathrm{argmin}_{W,H} ||M - W^{\intercal} HT||^2_F .
\end{align}
where $M=(A+A^2)/2$ is a normalized second-order adjacency matrix of the network, $W$ is a matrix of one-hot node embeddings and $H$ a feature transformation matrix. Final document embeddings are the concatenation of $W$ and $HT$.
Graph2Gauss (G2G) \cite{bojchevski2018deep} is an approach that embeds each node as a Gaussian distribution instead of a vector. The algorithm is trained by passing node attributes through a non-linear transformation via a deep neural network (encoder).      
GVNR-t \cite{brochier2019global} is a matrix factorization approach for document network embedding, inspired by GloVe \cite{pennington2014glove}, that simultaneously learns word, node and document representations. In practice, the following least-square objective is optimized: 
\begin{align} \label{eq:gvnrt}
\underset{U,W}{\mathrm{argmin}} \sum_{i=1}^{n_d} \sum_{j=1}^{n_d} \big(u_i \cdot \frac{\delta_j ~ W}{|\delta_j|_1} - \log (1 + x_{ij})\big)^2 .
\end{align}
where $x_{ij}$ is the number of co-occurrences of nodes $i$ and $j$, $u_i$ is a one-hot encoding of node $i$ and $\frac{\delta_j ~ W}{|\delta_j|_1}$ is the average of the word embeddings of document $j$.
Context-Aware Network Embedding (CANE) \cite{tu2017cane} consists in a mutual attention mechanism trained on a document network. It learns several embeddings for a document according to its different contextual documents, represented by its neighbors in the network. The attention mechanism selects meaningful features from text information in pairs of documents that explain their relatedness in the graph. A similar approach is presented in \cite{brochier2019link} where the links between pairs of documents are predicted by computing the mutual contribution of their word embeddings.     

In this work, we aim at constructing representations of documents that reflect their connections in a network. A key motivation behind our approach is to be able to predict a document's neighborhood given only its textual content. This allows our model to inductively produce embeddings for new documents for which no existing link is known. To that extend, Graph2Gauss is a similar approach. On the contrary, TADW and GVNR-t are not primarily designed for this purpose as they both learn one-hot embeddings for each node in the document network. Note that if some methods like GraphSage \cite{hamilton2017inductive}, SDNE \cite{wang2016structural} and GAE \cite{kipf2016variational} also enable induction on new nodes, they cannot deal with nodes that have no known connection. Also, our approach differs from CANE since this latter needs the neighbors of a document to generate its representation. \modelname{} learns to produce a single interpretable vector for each document in the network. 
In the next section, we review recent works in attention mechanisms for natural language processing (NLP) that inspired the conception of our method.   

\subsection{Attention Mechanism}

An attention mechanism uses a contextual representation to highlight or hide some parts of input data. Attention is an essential element of state-of-the-art neural machine translation (NMT) algorithms \cite{luong2015effective} by providing a powerful way to capture dependencies between words.   

The Transformer \cite{vaswani2017attention} introduces a formalism of attention mechanisms for NMT. Given a query vector $q$, a set of key vectors $K$ and a set of value vectors $V$, an attention vector is produced with the following formula: 
\begin{align} \label{eq:transformer}
v_a = \omega(qK^T)V .
\end{align}
$qK^T$ measures the similarity between the query and each key $k$ of $K$. $\omega$ is a normalization function such that all attention weights are positive and sum to 1. $v_a$ is then the weighted sum of the values $V$ according to the attention weights. Multiple attention vectors can be generated by using a set of queries $Q$. 

In CANE, as for various NLP tasks \cite{devlin2018bert}, an attention mechanism generates attention weights that represent the strengths of relation between pairs of input words. However, in this paper, we do not seek to learn dependencies between pairs of words, but rather between words and some global topics. In this direction, the Set Transformer \cite{lee2018set} constitutes a computationally efficient attention mechanism where the queries are replaced with a fixed-size set of learnable global inducing points. This model is originally not intended for NLP tasks, therefore we will explore the capacity of such inducing points to play the role of topic representations when applied to textual data. 

Even if we introduce the concept of topic vectors, the aim of this work is not to propose another topic model \cite{chang2009relational, srivastava2017autoencoding}. We hypothesize that the introduction of global topic vectors in an attention mechanism can (1) lead to useful representations of documents for different tasks and (2) bring an interpretable sight on the patterns learned by the model. Interpretability can help both machine learning practitioners to better refine their models and end users to understand automated recommendations.   
                                                        
\section{Method}\label{sec:method}  

We are interested in finding low dimensional vector space representations of a set of $n_d$ documents organized in a network, described by a document-term matrix $ X \in \mathbb{N}^{n_d \times n_w}$ and an adjacency matrix $A \in \mathbb{N}^{n_d \times n_d}$, where $n_w$ stands for the number of words in our vocabulary. The method we propose, Inductive Document Network Embedding (\modelname), learns to represent the words and topics underlying the corpus in a single vector space. The document representations are computed by combining words and topics through an attention mechanism. 

In the following, we first describe how to derive the document vectors from known word and topic vectors through a novel attention mechanism, the Topic-Word Attention (TWA). Next, we show how to estimate the word and topic vectors, guided by the links connecting the documents of the network.

\subsection{Representing Documents with Topic-Aware Attention}\label{TAA}

We assume a $p$-dimensional vector space in which both words and topics are represented. We note $W \in \mathbb{R}^{n_w \times p}$ the matrix that contain the $n_w$ word embedding vectors and $T \in \mathbb{R}^{n_t \times p}$ the matrix of $n_t$ topic vectors. Figure \ref{TAA} shows the matrix computation of the attention weights.

\paragraph{Topic-Word Attention} Given a document $i$ and its bag-of-word encoding $X_i \in \mathbb{N^+}^{n_w}$, we measure the attention weights between topics and words, $Z^i \in \mathbb{R}^{n_t \times n_w}$, as follows:

\begin{equation}
Z^i = \text{g}\big(TW^{\intercal}\text{diag}(X_i)\big).
\end{equation}
The activation function $g$ must satisfy two requirements: (1) all the weights are non-negative and (2) columns of $Z^i$ sum to one. The intuition behind the first requirement is that enforcing non-negativity should lead to sparse and interpretable topics. The second requirement transforms the raw weights into word-wise relative attention weights, which can be read as probabilities similarly to what is done in neural topic models \cite{srivastava2017autoencoding}. An obvious choice would be column-wise softmax, however, we empirically find that ReLU followed by a column-wise normalization performs best.

\paragraph{Document Representation}

Given $Z^i$, we are able to calculate topic-specific representations of the document $i$. From the perspective of topic $k$, the $p$-dimensional representation of document $i$ is:
\begin{equation} \label{eq:topic-doc}
D^i_k = \frac{Z^i_k \text{diag}(X_i)W}{|X_i|_1}.
\end{equation}

Similarly to Equation \ref{eq:transformer}, each topic vector, akin to a query, attends to the word vectors that play the role of keys to generate $Z^i$. The topic-specific representations are then the weighted sum of the values, also played by the word vectors. The final document vector is obtained by simple summation of all the topic-specific representations, which leads to $d^i = \sum_k{D^i_k}$. Scaling by $\frac{1}{|X_i|_1}$ in Equation \ref{eq:topic-doc} ensures that the document vectors have the same order of magnitude as the word vectors.

\begin{figure}
\begin{tikzpicture}[
        baseline=0cm,
        >={Stealth[length=3pt,width=6pt]},
        line width=1pt,
        Parenth/.style={
            left delimiter={(},
            right delimiter={)}
        },
        Matrix/.style={
            matrix of nodes,
            font=\scriptsize,
            align=center,
            column sep=1pt,
            row sep=1pt,
            nodes in empty cells,
        },
    ]

    \matrix[Matrix] at (0,0) (M1){ 
        1.2 & $\cdots$ & 2.2 \\
        \vdots  &  & \vdots \\
        0.1 & $\cdots$ & -1.4 \\
    };

    \matrix[Matrix,below=0.5 of M1] (M2){ 
        0.9 &$\cdots$ & -3.2 \\ 
        \vdots &  \vdots \\ 
        2.2 &$\cdots$  & 0.1 \\ 
    };

    \matrix[Matrix,left=1 of M2] (M3){ 
        -0.6& $\cdots$ & 1.2\\
        \vdots&  &\vdots \\
        1.3 & $\cdots$ & 2.4\\
    };
    
    \matrix[Matrix,right=1.5 of M2] (M4){ 
        0.15 &$\cdots$ & 0 \\ 
        \vdots &  & \vdots \\ 
        0.51 &$\cdots$  & 0.4 \\ 
    };

    \draw (M1-2-3.west)++(1.5,0) node[scale=2,transform shape]{${W^{i}}^\intercal$};
    \draw (M3-3-2.south)++(0,-0.5) node[scale=2,transform shape]{$T$};
    \draw (M4-3-2.south)++(0,-0.5) node[scale=2,transform shape]{$Z^i$};
    \draw (M2-3-2.south)++(0,-0.5) node[scale=1,transform shape, align=center]{Topic-Word dot \\products $T{W^{i}}^\intercal$};

    \node[draw,inner sep=0,fit=(M1-1-1)(M1-3-1)](HL1-M1){};
    \node[draw,inner sep=0,fit=(M1-1-3)(M1-3-3)](HL2-M1){};
    
    \node[draw,inner sep=0,fit=(M3-1-3)(M3-1-1)](HL1-M3){};
    \node[draw,inner sep=0,fit=(M3-3-3)(M3-3-1)](HL2-M3){};
    
    \node[draw,inner sep=0,fit=(M4-1-1)(M4-3-1)](HL1-M4){};
    \node[draw,inner sep=0,fit=(M4-1-3)(M4-3-3)](HL2-M4){};

    \node[Parenth,inner sep=0,fit=(M1)](BM1){};
    \node[Parenth,inner sep=0,fit=(M2)](BM2){};
    \node[Parenth,inner sep=0,fit=(M3)](BM3){};
    \node[Parenth,inner sep=0,fit=(M4)](BM4){};

    \draw[<->]
    (HL1-M1.north west)++(0,0.3) coordinate (temp) 
        -- (temp -| HL2-M1.east)
        node [midway,anchor=south]{$n_{w^i}$ words};
    \draw[<->]
    (HL1-M1.north west)++(-0.7,0) coordinate (temp) 
        -- (temp |- HL1-M1.south)
        node [midway,anchor=east]{$p$};
    
    \draw[<->]
    (HL1-M3.north west)++(0,0.3) coordinate (temp) 
        -- (temp -| HL2-M3.east)
        node [midway,anchor=south]{$p$};
    \draw[<->]
    (HL1-M3.north west)++(-0.7,0) coordinate (temp) 
        -- (temp |- HL2-M3.south)
        node [midway,anchor=east]{$n_t$ topics};
    
    \draw[<->]
    (HL2-M4.north east)++(0.7,0) coordinate (temp) 
        -- (temp |- HL2-M4.south)
        node [midway,anchor=west]{$n_t$};
    \draw[<->]
    (HL1-M4.north west)++(0,0.3) coordinate (temp) 
        -- (temp -| HL2-M4.east)
        node [midway,anchor=south]{$n_{w^i}$};
    
    \draw[->]
    (M2.east)++(-0.1,0) coordinate (temp) 
        -- (temp -| M4.west)
        node [midway,anchor=south, scale=2]{$g$};

 \end{tikzpicture}
 
 \caption{Matrix computation of the attention weights. Here $W^i$ is the compact view of $\text{diag}(X_i)W$ where zero-columns are removed since they do not impact on the result. $n_{w^i}$ denotes the number of distinct words in document $i$. Each element $z_{jk}$ of $Z^i$ is the column-normalized rectified scalar product between the topic vector $\boldsymbol t_{j}$ and the word embedding $ \boldsymbol {w^i}_{k}$ and represents the strength of association between the topic $j$ and the word $k$ in document $i$. The final document representation is then the sum of the topic-specific representations $D^i = \frac{Z^i W^{i}}{|X_i|_1}$.}
\label{TAA}
\end{figure}
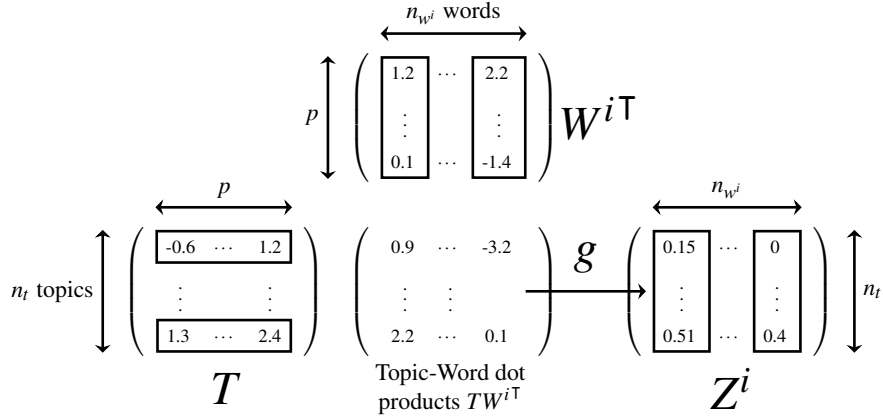

\subsection{Learning from the Network} \label{learning}

Since the corpus is organized in a network, we propose to estimate the parameters, $W$ and $T$, by leveraging the links between the documents. We posit that the representations of documents connected by a short path in the network should be more similar in the vector space than those that are far apart. Thus, we learn $W$ and $T$ in a supervised manner, through the training of a discriminative model.

Let $\Delta \in \mathbb\{0,1\}^{n_d \times n_d}$ be a binary matrix, so that $\delta_{ij} = 1$ if document $j$ is reachable from document $i$ and $\delta_{ij} = 0$ otherwise. We model the probability of a pair of documents to be connected, given their representations, in terms of the sigmoid of the dot-product of $d_i$ and $d_j$:

\begin{align}
P(Y = 1 | d_i, d_j ; W, T) = \sigma(d_i \cdot d_j).
\end{align}

Assuming the document representations are i.i.d, we can express the log-likelihood of $\Delta$ given $W$ and $T$:

\begin{align} \label{eq:log-likelihood}
\ell(W,T) &= \sum_{i=1}^{n_d} \sum_{j=1}^{n_d} \log P(Y = \delta_{ij} | d_i, d_j ; W,T) \nonumber \\
     &= \sum_{i=1}^{n_d} \sum_{j=1}^{n_d}  \delta_{ij} \log \sigma(d_i \cdot d_j) + (1 - \delta_{ij}) \log \sigma(-d_i \cdot d_j).
\end{align} 

Through the maximization of this log-likelihood via a first-order optimization technique, we back-propagate the gradient and thus learn the word and topic vectors that lead to the document representations that best reconstruct $\Delta$.

\section{Quantitative Evaluation} \label{sec:quanti-eval}

Common tasks in document network embedding are classification and link prediction. We assess the quality of the representations learned with {\modelname} for these tasks in two different settings: (1) a traditional setting where all links and documents are observed and (2) an inductive setting where only a fraction of the links and documents is observed during training. 

The first setting corresponds to a scenario where the goal is to propagate labels associated with a small portion of the documents. The second represents a scenario where we want to predict labels and links for new documents that have no network information, once the algorithm is already trained. This is common setting in real world applications. As an example, when a new user asks a new question on a Q\&A website, we would like to suggest tags for its question and to recommend potential similar questions. In this case, the only information available to the algorithm is the textual content of the question. 

\subsection{Experimental Setup}

We detail here the setup we use to train \modelname{}.

\paragraph{Computing the $\Delta$ matrix.} 

We consider paths of length up to 2 and compute the $\Delta$ matrix in the following manner:
\begin{equation}
\delta_{ij} = \begin{cases}
    		1 &\text{~if~} (A + A^2)_{ij} > 0,\\
    		0 &\text{~otherwise}.
    	\end{cases}
\end{equation}
This means that two documents are considered close in the network if they are direct neighbors or share at least one neighbor. Note that this matrix is the binarized version of the matrix TADW factorizes.   

\paragraph{Optimizing the log-likelihood.} We perform mini-batch SGD with the ADAM \cite{kingma2014adam} update rule. Because most document networks are sparse, rather than uniformly sampling entries of $\Delta$, we sample 5000 balanced mini-batches in order to favor convergence. We sample 16 positive examples ($\delta_{ij} = 1$) and 16 negative ones ($\delta_{ij} = 0$) per mini-bacth. Positive pairs of documents are drawn according to the number of paths of length 1 or 2 linking them. Negative samples are uniformly drawn. The impact of the number of steps is detailed in Section \ref{sec:hyper}.     

\subsection{Networks}

We consider 4 networks of documents of various nature:

\begin{itemize}
    \item A well-known scientific citation network extracted from Cora \footnote{https://linqs.soe.ucsc.edu/data}. Each document is an article labelled with a conference. 
    \item New York Times (NYT)  titles of articles from January 2007. Articles are linked according to common tags (\textit{e.g.} business, arts, technology) and are labeled with the section they appear in (\textit{e.g.} opinion, news). This network is particularly dense and documents have a short length.
    \item Two networks of the Q\&A website Stack Exchange (SE) \footnote{https://archive.org/details/stackexchange} from June 2019, namely \url{gaming.stackexchange.com} and \url{travel.stackexchange.com}. We only keep questions with at least 10 user votes and that have at least one answer with 10 user votes or more. We build the network by linking questions with their answers and by linking questions and answers of the same user. The labels are the tags associated with each question. 
\end{itemize}

\begin{table}[]
\center
\caption{General properties of the studied networks.}
\begin{tabular}{l|ccccccc}
                     & \# docs  & \# links   & \# labels & vocab size  & \# words per doc  & density & multi-label  \\ \hline
Cora                 &  2,211  &     4,771 &  7        &  4,333       & $67 \pm 32$   & 0.20\%   & no  \\
NYT                  &  5,135  & 3,050,513 &  4        &  5,748       & $24 \pm 17$   & 23.14\%  & no \\
Gaming               & 22,872  &   400,664 & 40        & 15,760       & $53 \pm 74$   & 0.15\%   & yes \\
Travel               & 15,087  &   465,696 & 60        & 14,539       & $70 \pm 73$   & 0.41\%   & yes \\
\end{tabular}
\label{tab:networks}
\end{table}

\subsection{Tasks and Evaluation Metrics}

For each network, we consider a traditional classification tasks, an inductive classification task and an inductive link prediction task. 

\begin{itemize}
    \item the traditional task refers to a setting where the model is trained on the entire network and the learned representations are used as features for a one-vs-all linear classifier with a training set of labelled documents ranging from 2\% to 10\% for multi-class networks and from 10\% to 50\% for multi-label networks.
    \item the inductive tasks refer to a setting where 10\% of the documents are removed from the network and the model is trained on the resulting sub-network. For the classification task, 
    a linear classifier is trained with the representations and the labels of the observed documents. Representations for hidden documents are then generated in an inductive manner, using their textual content only. Classifications and link predictions are then performed on these induced representations.  
\end{itemize}

To classify the learned representations, we use the LIBLINEAR \cite{fan2008liblinear} logistic regression \cite{kleinbaum2002logistic} algorithm and we cross validate the regularization parameter for each dataset and each model. Every experiment is repeated 10 times and we report the micro average of the area under the ROC curve (AUC). The AUC uses the probabilities of the logistic regression for all classes and evaluates the quality of the resulting ranking given the true labels. This metric is thus suitable for information retrieval tasks where we want to penalize wrong predictions depending on their ranks.
For link prediction, we rank pairs of documents according to the cosine similarity between their representations.

\subsection{Compared Representations}

For all document networks, we process the documents by tokenizing text into words, discarding punctuation, stop words and words that appear less than 5 times or in more than 25\% of the documents. We create document-term matrices that are used as input for 6 algorithms. Our baselines are representative of the different approaches for document NE. TADW and GVNR-t are based on matrix factorization whereas CANE and G2G are deep learning models. For each of them, we used the implementations of the authors:
\begin{itemize}
    \item LSA: we use a 256-dimensional SVD decomposition of the tf-idf vectors as a text-only baseline;
    \item TADW: we follow the guidelines of the original paper by using 20 iterations and a penalty term $\lambda=0.2$. For induction, we generate a document vector by computing the textual component $HT$ in Equation \ref{eq:tadw};
    \item Graph2gauss (G2G): we make sure the loss function converges before the maximum number of iterations;
    \item GVNR-t: we use $\gamma=10$ random walks of length $t=40$, a sliding window of size $l=5$ and a threshold $x_{\text{min}}=5$ with 1 iteration. For induction, we compute $ \frac{\delta_j ~ W}{|\delta_j|_1}$  in Equation \ref{eq:gvnrt};
    \item CANE: we use the same parameters as in the original paper;
    \item \modelname: we run all experiments with $n_t=32$ topic vectors. The effect of $n_t$ is discussed in Section \ref{sec:hyper}.
\end{itemize}

\subsection{Results Analysis}

Tables \ref{table:classif_cora_nyt} and \ref{table:classif_se} detail the AUC scores on the traditional classification task. We report the results for CANE only for Cora since the algorithm did not terminate within 10 hours for the other networks. In comparison, our method takes about 5 minutes to run on each network on a regular laptop. The classifier performs well on the representations we learned, achieving similar or better results than the baseline algorithms on Cora, Gaming and Travel Stack Exchange. However, regarding the New York Times network, GVNR-t and TADW have a slight advantage. Because of its high density, the links in this network are little informative which may explain the relative good scores of the LSA representations. We hypothesize that (1) TADW benefits from its input LSA features and that (2) GVNR-t benefits both from its random walk based matrix of node co-occurrences \cite{page1999pagerank}, which captures more precisely the proximities of the nodes in such dense network, and from the short length of the documents making the word embedding averaging efficient \cite{le2014distributed, arora2016simple}.

Table \ref{table:induction} shows the AUC scores in the inductive settings. For link prediction \modelname{} performs best on three networks, showing its capacity to learn meaningful word and topic representations according to the network structure. For classification, LSA and GVNR-t achieve the best results while \modelname{} reaches similar but slightly lower scores on all datasets. On the contrary, TADW and Graph2gauss show weaknesses on NYT and Gaming SE.

In summary, \modelname{} shows constant performances across all settings where other methods lack of robustness against the type of network or the type of task. A surprising result is the good scores of GVNR-t for inductive classification which we didn't expect given that its textual component only is used for this setting. However, for the traditional classification, GVNR-t has difficulties to handle networks wih longer documents. \modelname{} does not suffer the same problem because TWA carefully select discriminative words before averaging them. In Section \ref{sec:quali-eval}, we further show that \modelname{} learns meaningful representations of words and topics and builds interpretable document representations.    

\subsection{Impact of the Number of Topics and Convergence Speed} \label{sec:hyper}

\begin{figure}[]
\center
\begin{tikzpicture}
\begin{groupplot}[group style={group size=2 by 1},height=0.3\textwidth,width=0.5\textwidth]
    \nextgroupplot[ylabel={$AUC$}, xlabel={Number of topics $n_t$}, legend style={legend columns=-1, draw=none, fill opacity=0.5, draw opacity=1, fill=white, text opacity=1} ,legend to name=main]
    \foreach \column in {1,...,5}{\addplot+[] table [col sep=comma, x={topics}, y={\column}] {topics.csv};}
    \addlegendentry{2\%};\addlegendentry{4\%};\addlegendentry{6\%};\addlegendentry{8\%};\addlegendentry{10\%};
    \coordinate (top) at (rel axis cs:0,1);
    \nextgroupplot[ylabel={$AUC$}, xlabel={Number of steps}, xmode=log]
    \foreach \column in {1,...,4}{\addplot+[] table [col sep=comma, x={iter}, y={\column}] {iter.csv};}
    \coordinate (bot) at (rel axis cs:1,0);
  \end{groupplot}
\node[above=2em, right=-8.25em, inner sep=0pt] at  (top -| current bounding box.north) {\pgfplotslegendfromname{main}};
\end{tikzpicture}
\caption{Impact of the number of topics and of the number of steps on the traditional classification task on Cora with \modelname{}.}
\label{fig:hyper}
\end{figure}
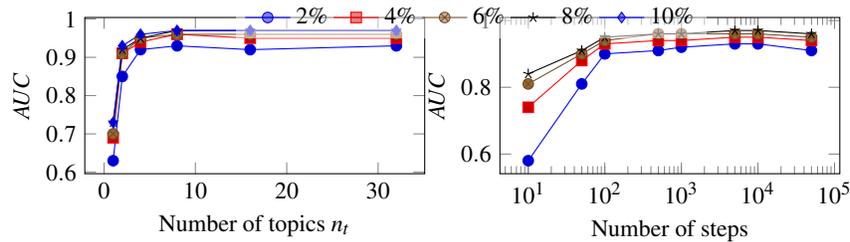

Figure \ref{fig:hyper} shows the impact of the number of topic vectors $n_t$ and of the number of steps (mini-batches) on the AUC scores obtained in traditional classification with Cora. Note that we observe a similar behavior on the other networks. We see that the scores improve from 1 to 16 topics and tend to stagnate for upper values. In a similar manner, performances improve up to 5000 iterations after which no increase is observed.  

\setlength{\tabcolsep}{4pt}

\begin{table}[]
\center
\caption{Micro AUC scores on Cora and NYT}
\label{table:classif_cora_nyt}
\begin{tabular}{l|ccccc|ccccc}
              & \multicolumn{5}{c|}{Cora}  & \multicolumn{5}{c}{NYT}       \\
              & 2\%      & 4\%      & 6\%      & 8\%      & 10\%  & 2\%      & 4\%      & 6\%      & 8\%      & 10\%  \\ \hline
LSA           & 67.54    & 81.76    & 88.63    & 89.68    & 91.43  & 79.90    & 82.06    & 81.18    & 83.99    & 86.06 \\
TADW          & 65.17    & 74.11    & 80.27    & 83.04    & 86.56  & 85.28    & \textbf{88.91}    & 87.49    & 89.39    & 88.72 \\
G2G           & 91.12    & 92.38    & 91.98    & 93.79    & 94.09  & 79.74    & 81.41    & 80.91    & 82.37    & 81.42 \\
CANE  & \textbf{94.40}    & \textbf{95.86}    & 95.90    & 96.37   & 95.88 & NA & NA & NA & NA & NA \\
GVNR-t        & 87.13    & 92.54    & 94.37    & 95.21    & 95.83  & \textbf{85.83}   & 87.67    & \textbf{88.76}    & \textbf{90.39}    & \textbf{89.90} \\
\modelname    & 93.34   & 94.93    & \textbf{95.98}    & \textbf{96.77}    & \textbf{96.68} & 82.40    & 84.60    & 86.16    & 86.72    & 87.98
\end{tabular}
\end{table}

\begin{table}[]
\center
\caption{Micro AUC scores on Stack Exchange Networks}
\label{table:classif_se}
\begin{tabular}{l|ccccc|ccccc}
              & \multicolumn{5}{c|}{Gaming}  & \multicolumn{5}{c}{Travel}       \\
              & 10\%      & 20\%      & 30\%      & 40\%      & 50\%  & 10\%      & 20\%      & 30\%      & 40\%      & 50\%  \\ \hline
LSA     & 86.73    & 88.51    & 89.51    & 90.25    & 90.18 & 80.18    & 83.77    & 83.40    & 84.12    & 84.60 \\
TADW    & 88.05    & 90.34    & 91.64    & 93.18    & 93.29 & 78.69    & 84.33     & 85.05     & 83.60    & 84.62 \\
G2G     & 82.12    & 84.42    & 85.14    & 86.10    & 87.84 & 66.04     & 67.48     & 69.67     & 70.94    & 71.58 \\
GVNR-t  & 89.09    & 92.60    & 94.14    & \textbf{94.79}    & 95.24 & 79.47     & 83.47     & 85.06     & 85.85    & 86.58 \\
\modelname   & \textbf{92.75}    & \textbf{93.53}    & \textbf{94.72}    & 94.61    & \textbf{95.57}  & \textbf{86.83}     & \textbf{88.86}     & \textbf{89.24}     & \textbf{89.31}    & \textbf{89.26}
\end{tabular}
\end{table}

\begin{table}[]
\center
\caption{Micro AUC scores for inductive classification and inductive link prediction}
\label{table:induction}
\begin{tabular}{l|cccc|cccc}
        & \multicolumn{4}{c|}{Inductive classification}  & \multicolumn{4}{c}{Inductive Link Prediction}\\
                & Cora  &NYT & Gaming   & Travel   & Cora  &NYT & Gaming   & Travel       \\ \hline
LSA             & 97.02           & \textbf{89.45}  & 90.70           & 85.88                   & 88.10           & 60.71           & 58.99           & 58.97  \\  
TADW            & 96.23           & 86.06           & 93.16           & 91.35                   & 84.82           & 69.10           & 57.00            & 57.91     \\
G2G             & 94.04           & 85.44           & 89.81           & 80.71                   & 81.58           & 74.22           & 58.18           & \textbf{59.50}     \\
GVNR-t          & \textbf{97.60}  & 88.47           & \textbf{96.09}  & \textbf{91.54}          & 82.27           & 71.15           & 59.71           & 58.39           \\
\modelname      & 96.58           & 88.21           & 95.22           & 90.78                   & \textbf{91.66}  & \textbf{77.90}  & \textbf{62.82}  & 58.43      
\end{tabular}
\end{table}

\section{Qualitative Evaluation} \label{sec:quali-eval}

We first show in Section \ref{sec:wordtopic} that \modelname{} is capable of learning meaningful word and topic vectors. Then, we provide visualizations of documents that highlight the ability of the topic-word attention to reveal topics of interest. For all experiments, we set the number of topics to $n_t = 6$.    

\subsection{Word and topic vectors} \label{sec:wordtopic}

Table \ref{table:topics_words} shows the closest words to each topic, computed as the dot product between their respective vectors, learned on Cora. Word and topic vectors are trained to predict the proximity of the nodes in a network, meaningless words are thus always dissimilar to the topic vectors, since they do not help to predict a link. This can be verified by observing the words that have the largest and the smallest norms, also reported in Table \ref{table:topics_words}. Even though the topics are learned in an unsupervised manner, we notice that, when we set the number of topics close to the number of classes, each topic seems to capture the semantics of one particular class.    

\begin{table}[]
\center
\caption{Topics with their closest words produced by \modelname{} on Cora and words whose vector $L_2$ norms are the largest (resp. the smallest) reported in parenthesis. The labels in this dataset are: Case Based, Genetic Algorithms, Neural Networks, Probabilistic Methods, Reinforcement Learning, Rule Learning and Theory.}
\label{table:topics_words}
\begin{tabular}{lp{0.8\textwidth}}
 Topic 1 &casebased, reasoning, reinforcement, knowledge, system, learning, decision\\
 Topic 2 &chain, belief, probabilistic, length, inference, distributions, markov\\ 
 Topic 3 &search, ilp, problem, optimal, algorithms, heuristic, decision\\ 
 Topic 4 &genetic, algorithm, fitness, evolutionary, population, algorithms, trees\\ 
 Topic 5 &bayesian, statistical, error, data, linear, accuracy, distribution\\ 
 Topic 6 &accuracy, induction, classification, features, feature, domains, inductive\\ \hline
Largest & genetic (8.80), network(8.07), neural(7.43), networks (6.94), reasoning (6.16)\\
Smallest & calculus (0.34), instability (0.34), acquiring (0.34), tested (0.34), le (0.34)\\
\end{tabular}
\end{table}

\subsection{Topic Attention Weights Visualization}

\begin{figure}[]
\centering
\fbox{\includegraphics[scale=0.35]{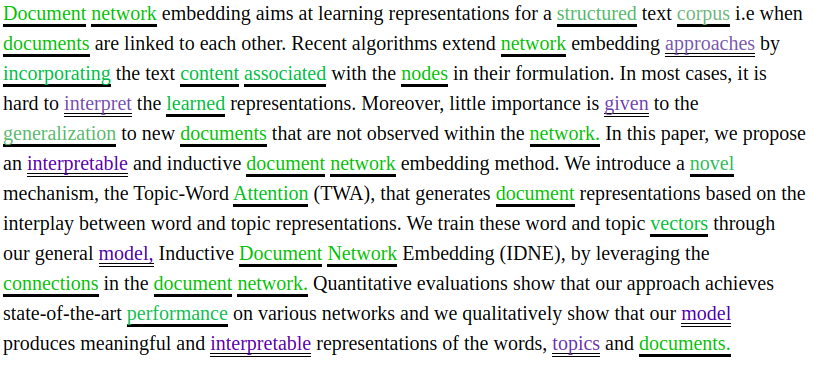}}
\caption{Topics provided by \modelname{} in the abstract of this very paper trained on Cora.}
\label{fig:topics_docs}
\end{figure}

To further highlight the ability of our model to bring interpretability, we show in Figure \ref{fig:topics_docs} the topics that most likely generated the words of a document according to TWA. The document is the abstract of this paper whose weights are inductively calculated with \modelname{} previously trained on Cora. We compute its attention weights $Z^i$ and associate each word $k$ to the maximum value of its column $Z^i_k$. We then colorize and underline each word associated to the two most represented topics in the document, if its weight is higher than $\frac{1}{2}$. We see that the major topic (green and single underline), that accounts for 32\% of the weights, deals with the type of data, here document networks. The second topic (blue and double underline), which represents 18\% of the weights, relates to text modeling, with words like \textquote{interpretable} and \textquote{topics}. 

\section{Discussion and Future Work} \label{sec:conclusion}

In this paper, we presented \modelname{}, an inductive document network embedding algorithm that learns word and latent topic representations via TWA, a topic-word attention mechanism able to produce interpretable document representations. We showed that IDNE performs state-of-the-art results on various network in different settings. Moreover, we showed that our attention mechanism provides an efficient way of interpreting the learned representations. In future work, we would like to study the effect of the sampling of the documents on the learned topics. In particular, the matrix $\Delta$ could capture other types of similarities between documents such as SimRank \cite{jeh2002simrank} which measures structural relatedness between nodes instead of proximities. This could reveal complementary topics underlying a document network and could provide interpretable explanations of the roles played by documents in networks. 


%

\clearpage

\bibliographystyle{splncs04}
\bibliography{bibliography}

\end{document}